\documentclass[letterpaper]{article} 
\usepackage{aaai25}  
\usepackage{times}  
\usepackage{helvet}  
\usepackage{courier}  
\usepackage[hyphens]{url}  
\usepackage{graphicx} 
\usepackage{natbib}  
\usepackage{caption} 
\usepackage{algorithm}
\usepackage{algorithmic}

\usepackage{multirow}
\usepackage{subfig}

\usepackage{amsfonts,amssymb}
\usepackage{newfloat}
\usepackage{listings}
\DeclareCaptionStyle{ruled}{labelfont=normalfont,labelsep=colon,strut=off} 
\floatstyle{ruled}
\newfloat{listing}{tb}{lst}{}
\floatname{listing}{Listing}
\nocopyright 

\title{SAM-SP: Self-Prompting Makes SAM Great Again}
\author{
    Chunpeng Zhou\equalcontrib, Kangjie Ning\equalcontrib, Qianqian Shen, Sheng Zhou, Zhi Yu, Haishuai Wang\thanks{Corresponding Author} \\
}
\affiliations{
    Zhejiang University, China \\
    {zhoucp@zju.edu.cn, haishuai.wang@gmail.com}

}

\usepackage{bibentry}

\begin{document}

\maketitle

\begin{abstract}
 The recently introduced Segment Anything Model (SAM), a Visual Foundation Model (VFM), has demonstrated impressive   capabilities in zero-shot segmentation tasks across diverse natural image datasets. Despite its success,  SAM encounters noticeably performance degradation when applied to specific domains, such as medical images. Current efforts to address this issue have involved fine-tuning strategies, intended to bolster the generalizability of the vanilla SAM.  However,  these approaches still predominantly necessitate the utilization  of domain specific expert-level prompts during the evaluation phase, which severely constrains the model's practicality. 
  To overcome this limitation, we introduce a novel self-prompting based fine-tuning approach, called SAM-SP, tailored for extending the vanilla SAM model. Specifically, SAM-SP leverages the output from the previous iteration of the model itself as prompts to guide subsequent iteration of the model. This self-prompting module endeavors to learn how to generate useful prompts autonomously and alleviates the dependence on expert prompts  during the evaluation phase, significantly broadening SAM's applicability. Additionally, we integrate a self-distillation module to enhance the self-prompting process further. Extensive experiments across various domain specific datasets validate the effectiveness of the proposed SAM-SP. Our SAM-SP not only alleviates  the reliance on expert prompts but also exhibits superior segmentation performance comparing to the state-of-the-art task-specific segmentation approaches, the vanilla SAM, and SAM-based approaches.
\end{abstract}

%

\section{Introduction}
Recent years have witnessed the remarkable evolution of Foundation Models, deemed as a generalized AI paradigm \cite{bommasani2021opportunities}.  These Foundation models have demonstrated exceptional performance across various real-world downstream tasks with minimal human intervention. Examples include the renowned ChatGPT \cite{ouyang2022training}, GPT-4 \cite{achiam2023gpt}, CLIP \cite{radford2021learning}, and BLIP \cite{li2022blip}, etc., all of which achieved remarkable success with minimal human intervention. 
However, these models were not specifically engineered for image segmentation \cite{minaee2021image}, an important computer vision task.
The recent introduced Segment Anything Model \cite{kirillov2023segment} (SAM),  a Foundation Visual Model for image segmentation, aims to address a range of downstream segmentation  tasks. Trained  on the extensive SA-1B  dataset \cite{kirillov2023segment}, which encompasses over 1 billion masks from 11 million images, SAM generates high-quality object masks and has demonstrated distinguished capabilities across diverse natural image datasets. Notably, SAM supports the flexible prompts from users, including point, box, and mask prompts. The zero-shot performance of SAM is impressive, often competitive with or even superior to prior fully supervised segmentation approaches \cite{kirillov2023segment} across various natural datasets.  

\begin{figure}[t]
    \centering
    \vspace{0.2cm}
    \includegraphics[width=7cm]{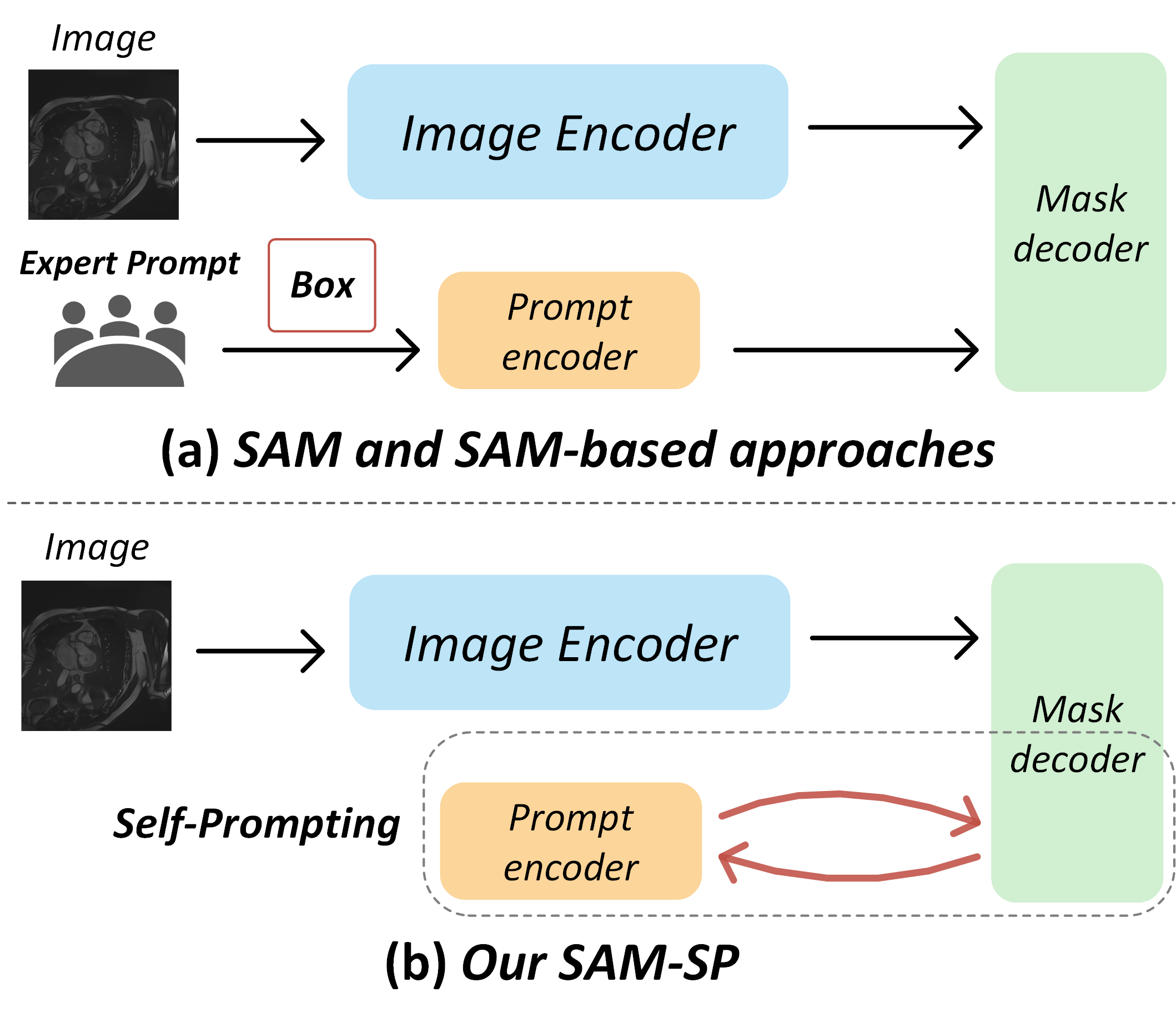}
    \caption{The illustration of our model. (a) SAM and SAM-based approaches both rely on the expert prompts during inference. (b) Our SAM-SP build a self-prompting module and do not rely on the expert prompts during inference.}
    \label{fig:1}
\end{figure}

Given these advancements, there is a natural inclination to apply SAM  to specific domains, such as medical images \cite{Litjens2017ASO,Hesamian2019DeepLT}, remote sensing images \cite{Jiang2022ASO,Shafique2022DeepLC}, and road damage images \cite{Yang2019RoadDA,Fan2019RoadCD}, etc. For instance,  precise medical image segmentation is crucial  for  various clinical applications, including disease diagnosis, treatment planning, and disease progression monitoring.
However, significant domain gaps exist between natural and medical images, including blurred boundaries and low contrast in medical images. Experimental results from previous studies \cite{ji2023segment, he2023computervision,zhou2023can,ma2023segment, mattjie2023zero, ning2023segment, ahmadi2023application} have indicates that directly applying SAM to these domains usually leads to noticeable performance degradation, with SAM significantly underperforming compared to traditional task-specific medical image segmentation approaches like U-net \cite{ronneberger2015u} and U-net++ \cite{8932614}.


To address these challenges and broaden SAM's  application range, recent approaches \cite{ma2023segment, wu2023medical} attempt to fine-tuning the vanilla SAM model with domain-specific data. For example, MedSAM \cite{ma2023segment} utilizes a simple fine-tuning method on a medical image dataset to adapt SAM to medical image segmentation tasks, while SAM-Med2D \cite{cheng2023sam} employs adaptation techniques \cite{chen2022adaptformer} on the image encoder of SAM. 
Despite these improved performance in specific domains compared to the vanilla SAM, they still heavily rely on the precision of provided prompts during inference \cite{wu2023medical} as shown in Fig \ref{fig:1}. In fact, their performance may degrade to the level of the vanilla SAM if the provided prompts are not sufficiently accurate \cite{wu2023medical}. Unfortunately, in contrast to natural images, the availability of precise prompts is difficult for specific domains, such as medical images, which heavily rely on expert domain knowledge. This limitation may hinder the widespread application of SAM. 
To abbreviate the need for expert prompts, we pose a question: \textit{can we make SAM perform well without any user prompts during inference?} 
Inspired by the  training  strategy of SAM, where SA-1B (the training dataset of SAM) only includes automatically generated masks by SAM itself as ground truth \cite{kirillov2023segment}, we attempt to produce reliable prompts independently without relying on expert prompts during inference. 

Consequently, we introduce a SAM-based segmentation framework with a novel proposed Self-Prompting strategy, called \textbf{SAM-SP}, allowing the model to produce prompts autonomously, eliminating the need for user prompts during inference.  SAM-SP builds upon the basic framework of the vanilla SAM, which encodes the input image using a ViT \cite{dosovitskiy2020image} based image encoder. Furthermore, we employ a low-rank-based finetuning strategy (LoRA) \cite{hu2021lora}, which allows SAM to better adapt to downstream tasks and makes training more efficient compared to full fine-tuning. And then the mask decoder utilize the image embeddings obtained by the image encoder to produce initial predictions. To alleviate the dependence on expert prompts during inference, we incorporate  a Self-Prompting module to generate prompts by the model itself. Specifically, the Self-Prompting module utilizes the first predictions to generate prompts for guiding the subsequent iteration of the model. In next iteration, the prompt encoder encodes the generated prompts into prompting embeddings, which are then fed to the mask decoder to obtain refined predictions. Though the supervised segmentation signals during training, SAM-SP learns to produce reliable and meaningful prompts by itself.  
Additionally, we utilize a segmentation based self-distillation module to further enhance the training of the self-prompting module. We leverage the later predictions as a teacher to guide previous predictions in our self-distillation module. SAM-SP gains mutual benefits from the self-distillation module. On the one hand, the final prediction provides additional supervision signals to guide the previous prediction of SAM-SP. On the other hand, the previous predictions provide more accurate prompts for subsequent predictions.  We conduct extensive experiments on diverse and publicly available datasets. The results indicate that  our proposed SAM-SP achieves satisfactory performance without user prompts, compared to state-of-the-art task-specific segmentation approaches, the vanilla SAM, and SAM-based approaches.

Our contributions in this paper are summarized as follows:
(\romannumeral1) We highlight the significance of diminishing reliance on expert prompts when deploying a visual foundation model in specific domains, such as the Segment Anything Model (SAM) in medical images.
(\romannumeral2) We introduce SAM-SP, a SAM-based framework which incorporates a novel self-prompting module, extending the vanilla SAM to explore its capability across various downstream tasks, without relying on expert prompts during inference. We further integrate a self-distillation module to enhance the self-prompting module.
(\romannumeral3) We build and release a novel and challenging segmentation dataset Seg-GPR for SAM, aiming for subgrade distress segmentation and collected by a 3-D ground penetrating radar.
(\romannumeral4) We conduct comprehensive evaluations of the proposed SAM-SP on  diverse and publicly available datasets, which demonstrate  superior performance without using any user prompts, compared to state-of-the-art task-specific segmentation approaches, the vanilla SAM, and SAM-based approaches.

\section{Related Work}

\subsection{Vision Foundation Model (LVM)}
Recent Large Language Model \cite{zhao2023survey} like ChatGPT \cite{ouyang2022training}, GPT-4 \cite{achiam2023gpt} have demonstrated exceptional performance in various NLP tasks. 
Similarly, Vision Foundation Models (LVM) have received huge attention.  For example,  CLIP \cite{radford2021learning} utilizes a contrastive learning strategy trained on  400 million image-text pairs collected from the internet, achieving excellent performance in a wide variety of downstream tasks.
Inspired by pre-training strategies in NLP \cite{kenton2019bert}, MAE \cite{he2022masked} employs an asymmetric encoder-decoder structure, aiming to reconstruct original images from random subsets of patches. Experimental results of MAE demonstrate the excellent transfer performance in downstream tasks. 
In another line,  DALL·E \cite{ramesh2021zero} is built based on a large autoregressive transformer, designed to generate images from textual descriptions.  
GPT-4 with Vision (GPT-4V) extends the capabilities of GPT-4 by incorporating visual modalities, enabling users to analyze image inputs \cite{achiam2023gpt}.
Distinct from previous VFMs, Segment Anything Model \cite{kirillov2023segment} (SAM) is the first VFM, specifically designed for segmentation tasks.  Trained on the large SA-1B dataset, SAM demonstrates exceptional zero-shot segmentation performance in natural images.

\subsection{SAM and SAM-based approaches}

To extend SAM's applicability to specialized domains, recent approaches have aimed to improve the vanilla SAM.
Notably, MedSAM \cite{ma2023segment} is the first attempt to extend SAM to medical images, employing a simple fine-tuning method with a large collected medical image dataset.   
SAM-Med2D \cite{cheng2023sam} and SAM-Adapter \cite{chen2023sam1} both utilize adaptation techniques \cite{chen2022adaptformer} on the image encoder of SAM, instead of the simple fine-tuning. 
The Medical SAM Adapter \cite{wu2023medical} (Med-SA) introduces Space-Depth Transpose (SD-Trans) and  Hyper-Prompting Adapter (HyP-Adpt) modules to achieve prompt-conditioned adaptation for medical image segmentation. 
SAMed \cite{zhang2023customized} applies a LoRA-based  \cite{hu2021lora} fine-tuning strategy to SAM's image encoder, jointly fine-tuning it together with the prompt encoder and mask decoder of SAM without user prompts. 
SAMUS \cite{lin2023samus} extends SAM by incorporating a parallel CNN branch, aiming to inject local features into the image encoder of SAM through cross-branch attention. The adapter technique \cite{chen2022adaptformer} is also used in SAMUS. 
Though these SAM-based approaches improve the generalization abilities of SAM, they predominantly ignore the difficulties  of precise prompts in specific domains, particularly in medical datasets. 
Consequently, we proposed a new framework SAM-SP which aims to produce prompts automatically during inference. 
Additionally, we highlight the difference between our approach and the related work by \citeauthor{wu2023self}, which introduces a simple logistic regression as the Self-Prompt Unit. The unit predicts a coarse mask from the image embeddings, and then uses it to obtain the bounding box as the prompt. In contrast, our SAM-SP learns to generate the prompts from the output of the SAM mask decoder, resulting in more precise prompts than those produced by simple logistic regression. Furthermore, we employ LoRA-based fine-tuning for better adaptation to downstream tasks, whereas only the logistic regression component is trainable in their framework \cite{wu2023self}, which may lead to limited generalization ability.

\section{Recap of the Segment Anything Model}
Before delving into the details of SAM-SP, we first review the basic framework of the first visual segmentation foundation 
model Segment Anything Model (SAM) \cite{kirillov2023segment}, containing three main components:  an image encoder, a prompt encoder, and a lightweight mask decoder. The image Encoder leverages an MAE \cite{he2022masked} pre-trained Vision Transformer (ViT) \cite{dosovitskiy2020image} to encoder the input images. The prompt encoder is designed to encoder the user-provided prompts, which supports multiple types of prompts, including both sparse (points, boxes) and dense (masks) prompts. 
Then, the mask decoder combines the image and prompt embeddings to predict the target mask. 
In this paper, we use $E(\ )$, $P(\ )$ and $M(\ )$ to denote image encoder, prompt encoder, and mask decoder, respectively. As discussed above, the vanilla  SAM struggles in specific domains and relies on the domain specific expert-level prompts during inference.

\begin{figure*}[t]
     \centering
    \includegraphics[width=17cm]{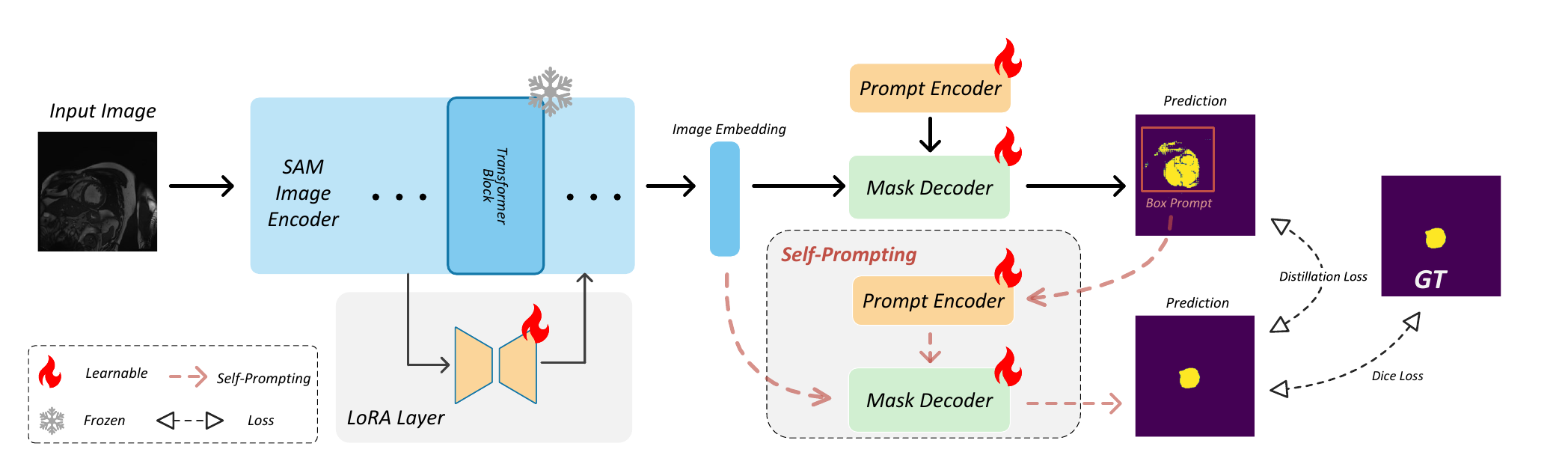}
    \caption{ The overall training architecture of our proposed SAM-SP, which inherits from SAM and contains three additional moudles:  LoRA-based fine tuning, Self-Prompting module and Self-Distillation module. our proposed SAM-SP significantly enhance the segmentation capability of SAM in specific domains and alleviates  the reliance on expert prompts during inference.
    The two prompt encoders here are shared with parameters, the same with two mask decoders.}
    \label{fig:2}
\end{figure*}

\section{Methodology}

\subsection{Overview of Our Method}
As illustrated  in Figure \ref{fig:2}, our proposed SAM-SP aims to enhance the segmentation capability of SAM in specific domains while reducing the reliance on expert prompts.   SAM-SP is an end-to-end framework built  on the vanilla SAM, containing three additional modules: LoRA-based fine-tuning, the Self-Prompt module and the Self-Distillation module.

\subsection{LoRA-based Fine-tuning}
As discussed above, significant domain gaps usually arise between downstream datasets (e.g. medical images) and natural images, for which the vanilla SAM is engineered. To mitigate these gaps and improve the generalization abilities of SAM, fine-tuning has been widely adopted in previous works \cite{ma2023segment,cheng2023sam,wu2023medical,zhang2023customized,lin2023samus}. 
Instead of fine-tuning the all parameters of SAM, we freeze the image encoder, and only fine-tune the prompt encoder and mask decoder to minimize computational costs, given that the image encoder dominates the majority (more than 95\%) computational overhead in SAM. 
To further enhance the quality of image embeddings from the image encoder, we employ a low-rank-based fine-tuning strategy (LoRA) \cite{hu2021lora}, which approximates the low rank update of the parameters in the image encoder. LoRA enables SAM to adapt effectively to downstream tasks, while maintaining training efficiency compared to full fine-tuning. Additionally, LoRA merges the trainable matrices with the frozen weights when deployed, introducing no additional inference latency. For a pre-trained weight matrix $\mathbf{W} \in \mathbb{R}^{d \times k} $ in the SAM image encoder, we suppose its update as follows: $\mathbf{\hat{W}} = \mathbf{W} +  \Delta \mathbf{W} =  \mathbf{W} + \mathbf{BA} $, where $\mathbf{\hat{W}} \in \mathbb{R}^{d \times k} $ denotes the updated weight matrix, and a low-rank decomposition $\Delta \mathbf{W} = \mathbf{BA}, \mathbf{A} \in \mathbb{R}^{r \times d}, \mathbf{B} \in \mathbb{R}^{d \times r }$ models this weight update. 
We set $4$ as the default rank of the low-rank decomposition in our implementation. 


\subsection{Self-Prompting Module}
As depicted in Fig \ref{fig:2}, we introduce a Plug-and-Play Self-Prompting module to alleviate the need of expert prompts. 
Previous works typically simulate user prompt according the expert annotations during the training stage.  For example, MedSAM \cite{ma2023segment}, Med-SA \cite{wu2023medical} and SAM-Med2D  \cite{cheng2023sam} all employ the ground truth prompts during both training and evaluation. 
However, this may lead to models becoming overly reliant on provided prompts.
To reduce this reliance and achieve training-test matching, we do not use any user prompts during both training and evaluation. The prompt encoder in SAM can work without any prompt input, and will update a default embedding during training \cite{kirillov2023segment}.
Formally, given an image $x$, we first obtain the image embedding $E(x)$ via the image encoder $E(\ )$ in SAM with LoRA. The vanilla prediction $\hat{y_0}$ of our model without prompts is given by:
\begin{equation}
    \hat{y_0} = M(E(x),  P(none) )
\end{equation}
To further alleviate the need of expert prompts, our model generates the prompts by itself without manual input. Specifically, we use the vanilla prediction $\hat{y_0}$ from the first iteration to generate a prompt to guide the subsequent iteration. 
And a box prompt is produced using the maximum and minimum coordinates of the predicted mask $\hat{y_0}$, and we denote this self-prompting process as $SP(\ )$. Then, the prompt $SP(\hat{y_0})$ produced by the model itself is  into the prompt encoder $P(\ )$, and the resulting self-prompt embedding $P(SP(\hat{y_0}))$ combined with the image embedding $E(x)$ is fed into the mask decoder $M(\ )$,  formulated as:
\begin{equation}
    \hat{y_1} = M(E(x),  P(SP(\hat{y_0})) )
\end{equation}
where $\hat{y_1}$ denotes the predicted result from the self-prompting process. This module enables the model to learn producing the prompts by itself, further reducing reliance on expert prompts.

Notably, the obtained image embeddings are used repeatedly during the self-prompting process, without adding obvious additional computational costs. And the introduced self-prompting module does not introduce any learnable parameters, keeping SAM-SP efficient.

\subsection{Self-Distillation Module}
Inspired by the recent progresses in Knowledge Distillation \cite{gou2021knowledge, wang2021knowledge}, we utilize a segmentation based self-distillation module to further enhance the training of the self-prompting process. 
As mentioned earlier, we obtained the final predicted mask $\hat{y_1}$ from the self-prompting process. Additionally, we also obtain the the vanilla predictions $\hat{y_0}$ from our model.
Specially, we employ the final prediction $\hat{y_1}$ as the teacher to guide the previous prediction $\hat{y_0}$ as the student. In this self-distillation module, SAM-SP gains mutual benefits from both the teacher and student. The final prediction provides additional supervisory signals to guide the previous prediction. Simultaneously, the previous predictions help to refine the subsequent predictions by providing more accurate prompts. The self-distillation process is formulated as: $L_{SD} = KL(\hat{y_0},\hat{y_1})$, 
where $L_{SD}$ denotes the knowledge distillation loss, with Kullback-Leibler Divergence as our default implementation. 

Notably, the self-distillation module also does not introduce any learnable parameters and is deprecated during evaluation, further keeping SAM-SP efficient.

\subsection{Training and Evaluation}
The overall training loss of SAM-SP is computed as:
\begin{equation}
    L = Dice(\hat{y_1},y) + \alpha L_{SD}
\end{equation}
where $y$ symbolizes the ground truth, $Dice(\ )$ denotes  Dice loss, $\alpha$ serves as the weighting factor. Notably, our SAM-SP does not use any user-provided prompts during both training and inference. 


\section{Experiments}

\begin{table*}[t] \small
\caption{Quantitative comparison on Polyp Segmentation of different approaches}
\begin{tabular}{l|cccccccccc}
\hline \multirow{2}{*}{ Models } & \multicolumn{2}{c}{ CVC-300 } & \multicolumn{2}{c}{ CVC-ClinicDB } & \multicolumn{2}{c}{ Kvasir } & \multicolumn{2}{c}{ ColonDB } & \multicolumn{2}{c}{ ETIS-LaribDB } \\
& DICE & $\mathrm{IoU}$ & DICE & $\mathrm{IoU}$ & DICE & $\mathrm{IoU}$ & DICE & $\mathrm{IoU}$ & DICE & $\mathrm{IoU}$ \\
\hline U-Net \cite{ronneberger2015u}  & 71.0 & 62.7 & 82.3 & 75.5 & 81.8 & 74.6 & 51.2 & 44.4 & 39.8 & 33.5 \\
UNet++ \cite{zhou2018unet++} & 70.7 & 62.4 & 79.4 & 72.9 & 82.1 & 74.3 & 48.3 & 41.0 & 40.1 & 34.4 \\
PraNet  \cite{fan2020pranet}& 87.1 & 79.7 & 89.9 & 84.9 & 89.8 & 84.0 & 71.2 & 64.0 & 62.8 & 56.7 \\
UACANet-L \cite{kim2021uacanet} & 88.21 & 80.84 & 91.07 & 86.7 & 90.83 & 85.95 & 72.57 & 65.41 & 63.89 & 56.87 \\
 SSFormerPVT  \cite{wang2022stepwise} & \textbf{89.46} & \textbf{82.68} & 92.88 & 88.27 & 91.11 & 86.01 & 79.34 & 70.63 & 78.03 & 70.1 \\
PolypPVT  \cite{dong2023polyp} & 88.71 & 81.89 & \textbf{93.08} & \textbf{88.28} & \textbf{91.23} & \textbf{86.30} & \textbf{80.75} & \textbf{71.85} & \textbf{78.67 }&\textbf{ 70.97} \\
 \hline 
SAM   \cite{kirillov2023segment}     & 45.00 & 38.62  & 33.29 & 25.64 & 61.48 & 53.74  & 29.33 & 31.23 & 24.76 & 20.37 \\
SAMed   \cite{zhang2023customized}  & 83.63 & 76.27 & 83.34 & 76.32 & 88.01 & 81.61 & 70.57 & 62.69  & 60.13 & 52.05 \\
SAM-Med2D  \cite{cheng2023sam}      & 84.81 & 77.49 & 85.91  & 80.60 & 87.06  & 80.88 & 69.08 & 60.79 & 59.80 & 53.00 \\
Med-SA    \cite{wu2023medical}      & 83.63  & 76.27  & 86.32 & 80.80 & 87.11 & 80.53 & 73.68 & 64.97 & 59.04 & 52.32 \\
\hline
\textbf{SAM-SP (Ours) }        & \textbf{88.94} & \textbf{82.55} & \textbf{85.91} & \textbf{80.37} &  \textbf{90.57} & \textbf{85.46} & \textbf{74.67} & \textbf{67.24 }& \textbf{64.87} & \textbf{58.11}\\

\hline
\end{tabular}
\label{tab:1}
\end{table*}

In this section, we aim to validate the performance of our proposed SAM-SP model without the use of user-provided prompts during inference. 
To comprehensively assess the effectiveness of SAM-SP compared to state-of-the-art approaches, we conduct extensive experiments across 10 publicly available datasets. 
\subsection{Benchmark Dataset}
\noindent \textbf{Medical Image Segmentation.} Medical image segmentation is a critical and challenging task.  In this evaluation, we select Polyp Segmentation \cite{jha2020kvasir} and Skin Lesion Segmentation \cite{berseth2017isic} tasks. For Polyp Segmentation, we use 5 datasets: Kvasir-SEG \cite{jha2020kvasir}, ClinicDB \cite{bernal2015wm}, ColonDB \cite{tajbakhsh2015automated}, Endoscene \cite{vazquez2017benchmark}, and ETIS-LaribDB \cite{silva2014toward}. Following previous settings \cite{fan2020pranet}, we adopt 900 and 548 images from the ClinicDB and Kvasir datasets as the training set, and the remaining 64 and 100 images are employed as the respective test sets. To evaluate the generalization performance, we test the model on three unseen datasets: EndoScene, ColonDB and ETIS. 
For skin lesion segmentation, two skin lesion datasets are used: ISIC 2017 \cite{berseth2017isic} with 2050 dermoscopy images, and ISIC 2018 \cite{codella2019skin} with 2694 dermoscopy images. To ensure a fair comparison, we follow the 7:3 train/test split strategy as outlined in \cite{ruan2023ege}.

\noindent \textbf{Remote Sensing Images.}  Remote sensing images \cite{zhang2020dense} typically cover a wide scope with complex backgrounds and diverse noise interference. 
ORSSD dataset \cite{li2019nested} contains 600 images for training and the rest 200 images for testing. EORSSD dataset \cite{zhang2020dense} is divided into two parts, where 1400 images for training and 600 images for testing. 

\noindent \textbf{Distress detection.}  Distress detection plays a crucial component in transportation infrastructure. We build a novel subgrade distress detection dataset collected by an 3-Dimension ground penetrating radar (GPR) which is wildly employed in infrastructure health monitoring \cite{kim2021novel,liu2021application,zhou2023multi}. This GPR dataset, named seg-GPR, consists of 395 images, with an 80/20 split for training and testing.

\subsection{Implementation Details and Baselines}
We compare our proposed SAM-SP to several baselines, including task-specific segmentation approaches, the vanilla SAM, and SAM-based approaches. 
We re-implemented all SAM-base approaches using their open-source code, including SAM-Med2D \cite{cheng2023sam}, Medical SAM Adapter (Med-SA) \cite{wu2023medical} and SAMed \cite{zhang2023customized}.
For a fair comparison, these approaches are trained on  the same training set of a dataset, and we uniformly use the ViT-B as the image encoder for SAM and SAM-based approaches. 
Our paper mainly focus the downstream performance without any use prompt. Consequently, no prompts are accessible for any approach during both training or inference periods, unless otherwise specified.  We employ the AdamW optimizer \cite{loshchilov2019decoupled} in all experiments. For the learning rates and weight decays, we both make a gird search in the set \{1e-2,1e-3,1e-4,1e-5\}. Similarly, we search for the optimal number of training epochs of different approaches in the range \{100, 200, 300\}.
Specially, we employ the warmup strategy \cite{he2016deep} to stabilize the training process. 
Due our proposed SAM-SP is cost-effectiveness, SAM-SP can be run on a single NVIDIA RTX 3090 GPU.
We use two wildly adopted evaluation metrics:  Dice Similarity Coefficient (DICE) and Intersection over Union (IoU), defined as: $Dice=\frac{2T P}{F P+ 2TP + F N} $ and $I O U=\frac{T P}{T P+F P+F N}$, respectively.



\begin{table}[t] \footnotesize
\caption{Quantitative comparison on Skin Lesion Segmentation of different approaches}
\setlength{\tabcolsep}{1mm}{
\begin{tabular}{l|cccc}
\hline 
\multirow{2}{*}{Methods} & \multicolumn{2}{c}{ISIC2017} & \multicolumn{2}{c}{ISIC2018} \\
& DICE   & IoU   & DICE   & IoU   \\
\hline
U-Net    & 86.99 & 76.98 & 87.55 & 77.86 \\
UNet++  \cite{zhou2018unet++}  &   -    &   -    & 87.83 & 78.31 \\
SANet  \cite{wei2021shallow}   &    -   &  -     & 88.59 & 79.52 \\
TransFuse\cite{zhang2021transfuse}                   & 88.40  & 79.21 & 89.27 & 80.63 \\
MALUNet \cite{ruan2022malunet}  & 88.13 & 78.78 & 89.04 & 80.25 \\
EGE-Unet \cite{ruan2023ege}  & \textbf{88.77} & 79.81 & 89.04 & 80.25 \\
\hline
SAM  & 53.28 & 40.80 & 58.79 & 46.06 \\
SAM-Med2D     & 87.01 & 79.63 & 88.92 & 81.87 \\
SAMed       & 85.86 & 77.51  & 88.73 & 79.84 \\
Med-SA          & 87.78  & 80.35& 88.75 & 81.77 \\
\hline
\textbf{SAM-SP (Ours)}     & \textbf{87.66} & \textbf{80.01} & \textbf{89.39} & \textbf{82.31} \\
\hline
\end{tabular}
}
\label{tab:2}
\end{table}

\begin{table}[t]\footnotesize
\caption{Quantitative comparison on Remote Sensing Images Segmentation of different approaches}
\centering
\begin{tabular}{l|cccc}
\hline
\multirow{2}{*}{Methods} & \multicolumn{2}{c}{ORSSD} & \multicolumn{2}{c}{EORSSD} \\
& DICE   & IoU   & DICE   & IoU   \\
DAFNet \cite{zhang2020dense}           & 87.4  & 82.3  & 83.0    & 80.0    \\
MJRBM \cite{tu2021orsi}             & 85.3  & 81.7  & 82.2  & 79.3  \\
RRNet \cite{cong2021rrnet}          & -     & -    & 86.2  & 83.4  \\
GateNet    \cite{zhao2023towards}           & 88.5  & 83.9  & 89.2  & 85.0   \\
\hline
SAM                   & 41.54  & 37.17  & 32.90  & 29.05  \\
SAM-Med2D            & 89.38  & 83.19 & 88.13 & 81.68 \\
SAMed               & 90.42 & 84.57 & 88.19 & 81.45 \\
Med-SA                  & 90.37 & 84.10  & 90.28 & 84.35 \\
\hline
\textbf{SAM-SP (Ours) }         & \textbf{92.66} & \textbf{87.53} & \textbf{91.01} & \textbf{85.22 }\\
\hline
\end{tabular}
\label{tab:3}
\end{table}


\begin{table*}[htbp]
\centering
\begin{minipage}{0.2\textwidth}
\centering \scriptsize
{\caption{Performance on  Distress detection. }}
\begin{tabular}{l|cc}
\hline
\multirow{2}{*}{Methods} & \multicolumn{2}{c}{Seg-GPR}                                          \\
                         & \multicolumn{1}{c}{DICE} & \multicolumn{1}{c}{IoU} \\
\hline
U-Net    & 66.23 & 54.87 \\
SAM      & 20.89 & 11.28  \\
SAMed      & 74.33 & 64.17 \\
SAM-Med2D   & 72.95 & 62.46  \\
\textbf{SAM-SP (Ours)}   & \textbf{77.85} & \textbf{67.90}  \\
\hline 
\end{tabular}
\label{tab:top}
\end{minipage}
\hfill
\begin{minipage}{0.75\textwidth}
\centering \footnotesize
\caption{Ablation Studies on SAM-SP}
\begin{tabular}{lllllllllll}
\hline
\multirow{2}{*}{Modules} & \multicolumn{2}{c}{ISIC2017} & \multicolumn{2}{c}{ISIC2018} & \multicolumn{2}{c}{ORSSD} &\multicolumn{2}{c}{Kvasir} & \multicolumn{2}{c}{Seg-GPR} \\
 & DICE   & IoU   & DICE   & IoU   & DICE   & IoU   & DICE  & IoU & DICE   & IoU   \\
 \hline
vanilla SAM       & 53.28  & 40.80 & 58.79 & 46.06 & 41.54  & 37.17  & 61.48 & 53.74 & 20.89 & 11.28 \\
+ LoRA            & 85.86 & 77.51  & 88.73 & 79.84 & 90.42 & 84.57  & 88.01 & 81.61 & 74.33&64.17\\
+ SP              & 87.03 & 79.69           & 89.17 & 81.96 & 92.25 & 86.99 & 89.95  & 84.56 & 77.32& 67.20\\
+ KD (Ours)       & 87.66 &  80.01           & 89.39 & 82.31 & 92.66 & 87.53 & 90.57 & 85.46 & 77.85 & 67.99\\
\hline     
\end{tabular}
\label{tab:ablation}
\end{minipage}
\end{table*}


\begin{figure}[t]
     \centering
    \includegraphics[width=8.2cm]{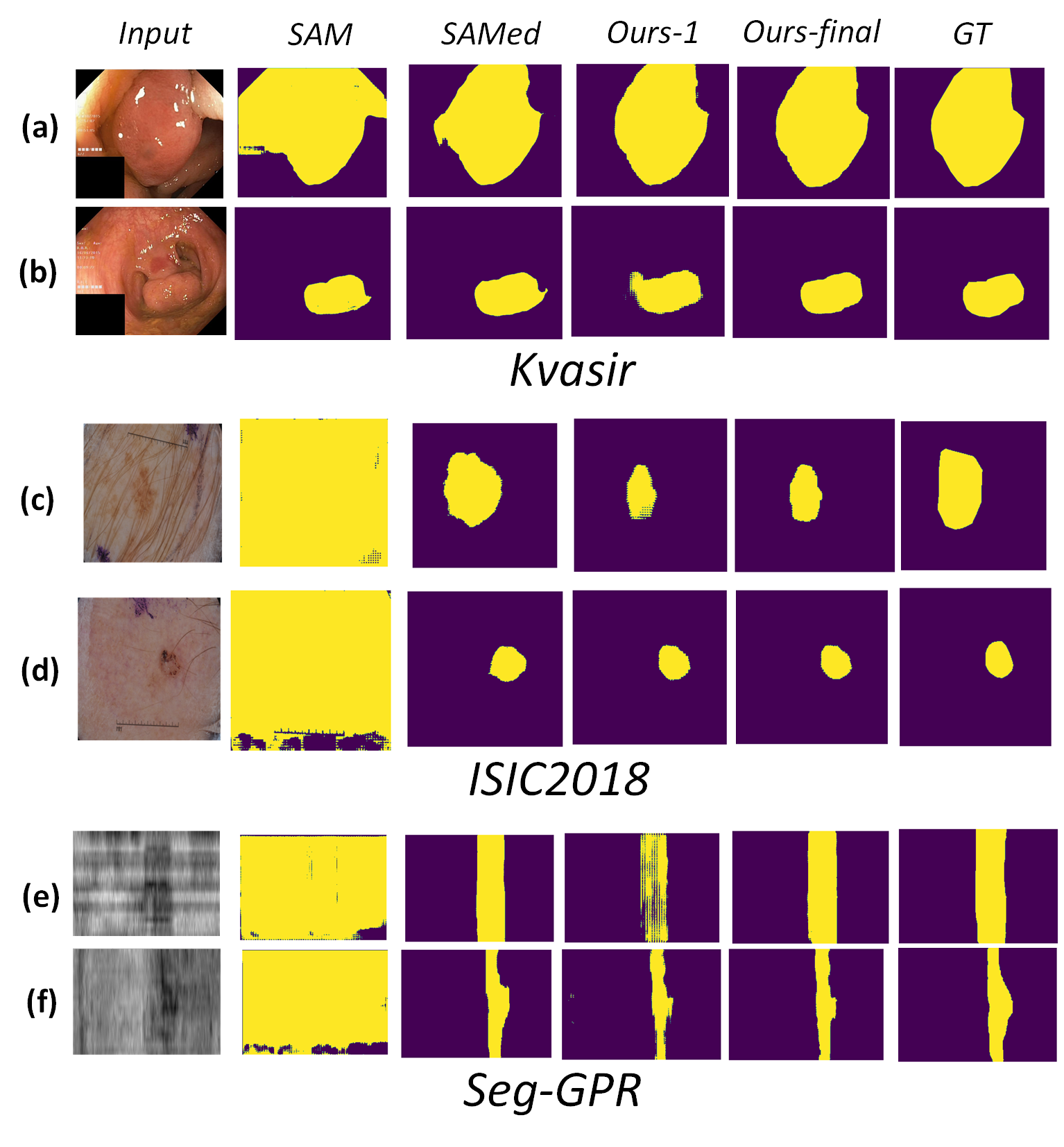}
    \caption{Visualization of segmentation results}
    \label{fig:vis}
\end{figure}

\subsection{Main Results}

\subsubsection{Quantitative results.}

In Table \ref{tab:1}, we summarize the qualitative results of our SAM-SP compared to other methods on the Polyp Segmentation task. Approaches in the top row are the task-specific segmentation approaches, and approaches in the middle row are SAM and SAM-based approaches. 
Specially, in our implementation, we provide user prompts for the vanilla SAM considering that it is a zero-shot Segmentation, following previous works
\cite{wu2023medical}. Here, SAM denotes the SAM with point prompts provided during inference, sampled within the ground-truth box. Similarly, Table \ref{tab:2}, \ref{tab:3} and 4 summarize the quantitative results of our SAM-SP with other methods on Skin lesion segmentation, Remote Sensing Images Segmentation, Distress segmentation respectively. 
Firstly, observed from these quantitative results, our proposed SAM-SP achieve distinguished performance compared to the vanilla SAM and a series of SAM-based approaches across most downstream segmentation tasks, even without any prompts provided. For example, SAM-SP surpasses SAMMed2D with an obvious improvement on CVC-300, Kvasir, ColonDB and ETIS-LaribDB datasets in the Polyp Segmentation task.  
Similarly, SAM-SP outperform SAMed with an obvious improvements on both the ISIC2018 and ISIC2017 datasets in the Skin lesion segmentation task. 
The similar performance improvements also be found on ORSSD, EORSSD and Seg-GPR Dataset.
These improvements can be attributed  to the our proposed self-prompting and self-distillation modules in SAM-SP, which enable the model to produce prompts autonomously, thus reducing reliance of user prompts during inference.
Secondly, unfortunately, the vanilla SAM struggles to segment these domain-specific targets, and its performance is significantly lower than other SAM-based approaches. Though we provide the ground-truth based point prompts for SAM during inference, SAM still can not achieve the satisfied performances.  Especially for our built Seg-GPR, it is a very challenging for the vanilla SAM, while our SAM-SP perform well in this dataset.
Thirdly, our quantitative results also indicate that task-specific segmentation approaches still perform very competitively and, in some cases, still better than SAM-based approaches. For instance, while our proposed SAM-SP surpasses the state-of-the-art task-specific segmentation approaches in Skin lesion segmentation and Remote Sensing Images Segmentation tasks, it does not outperform the SOTA Polyp Segmentation tasks, such as SSFormerPVT and PolypPVT, which are specifically designed for the Polyp Segmentation with sophisticated network structures. How to allow Vision Foundation Models surpass the all SOTA task-specific approaches deserves further investigation.

\subsubsection{Qualitative results.}

We showcase the qualitative results of our proposed SAM-SP in Figure \ref{fig:vis} compared to the vanilla SAM and a SAM-based method SAMed. The top two rows show samples from Polyp Segmentation dataset, the middle two rows from  the Skin lesion segmentation dataset, and the bottom two from the Seg-GPR dataset. 
Ours-1 denotes the first prediction of SAM-SP, and Ours-final means the final prediction of SAM-SP.
These results indicate that, due to the absence of fine-tuning, the zero-shot SAM can not produce reliable segmentation results. 
 Though SAMed achieves better segmentation results compared to the vanilla SAM, it still usually produce unclear boundary, as shown in \ref{fig:vis}(a). 
In contrast, our SAM-SP utilize the proposed self-prompting process to produce prompts autonomously, thus exhibiting more accurate predicted masks which closely resemble the ground truths in these samples from different datasets. 
Specially, we observe that our first output still performs competitively in some cases, which contributes to that our proposed self-distillation module allow later prediction guide previous prediction. 
In summary, our proposed SAM-SP consistently performs well and produce satisfactory and high-quality segmentation results across various datasets, without relying on the user prompts during inference.




\subsection{Ablation Studies}

As described earlier, our proposed SAM-SP inherits the basic structure of the vanilla SAM, and contains three additional modules: LoRA-based fine tuning, the self-prompting module, and the self-distillation module. Consequently, we investigate the effectiveness of each module in SAM-SP and summarize the experimental results in Table \ref{tab:ablation}. These ablation results indicate that LoRA-based fine tuning (denoted as LoRA) provides the first performance boost compared to the vanilla SAM. The further performance boost occurs when the self-prompting module (denoted as SP) is introduced, which underscores its effectiveness. 
Finally, the employment of self-distillation module (denoted as KD) brings the additional performance gains, based on previous modules. The self-distillation module allows the previous and later prediction of SAM-SP gain benefits mutually. In summary, by coupling these three modules, SAM-SP achieves the satisfactory segmentation performance across various downstream datasets without relying on user-provided prompts.

\subsection{Model Analysis}

\subsubsection{Self-Prompting Varieties}
\begin{figure} [t]
	\centering
	\subfloat[Kvasir]{
		\includegraphics[scale=0.5]{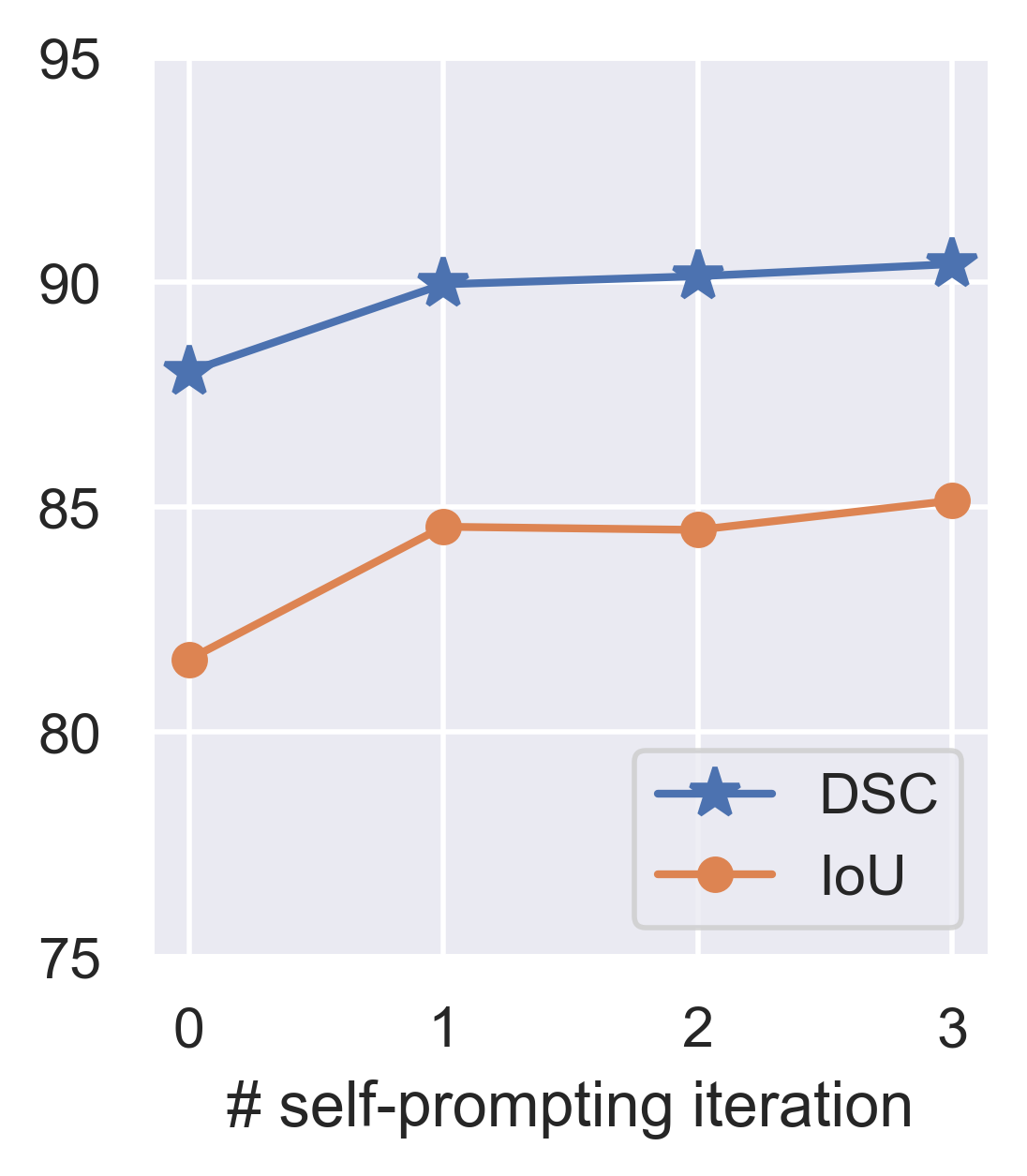} }
	\subfloat[Seg-GPR]{
		\includegraphics[scale=0.5]{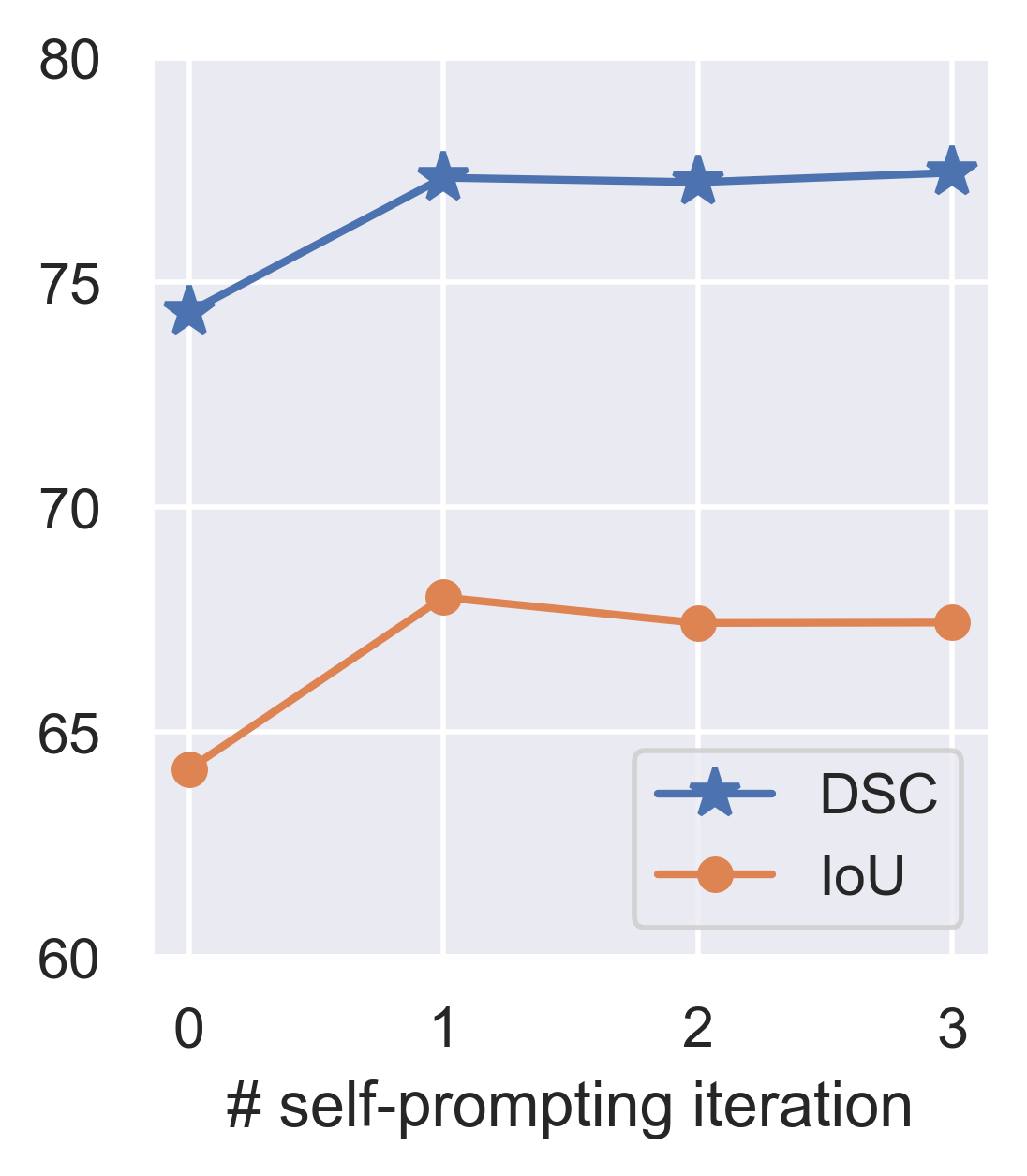}}

	\caption{Quantitative comparison with different number of self-prompting iterations.}
	\label{fig:SP} 
\end{figure}

In our implementation of SAM-SP, we typically conduct the self-prompting process only once by default. Consequently, we investigate the impacts with multiple  self-prompting processes in our proposed SAM-SP, and summaries experimental results with varying numbers of self-prompting processes in Figure \ref{fig:SP}. Firstly, observed from these results, the first self-prompting process brings the most significant performance boost compared to the second and third. Secondly, the more times of self-prompting processes may yield diminishing returns which may due to the convergence  of output results in our self-prompting module. Therefore, for a balance between performance and efficiency, we default to a single self-prompting process in SAM-SP.


\subsubsection{Training Varieties}
\begin{figure}[t]
     \centering
    \includegraphics[width=7.5cm]{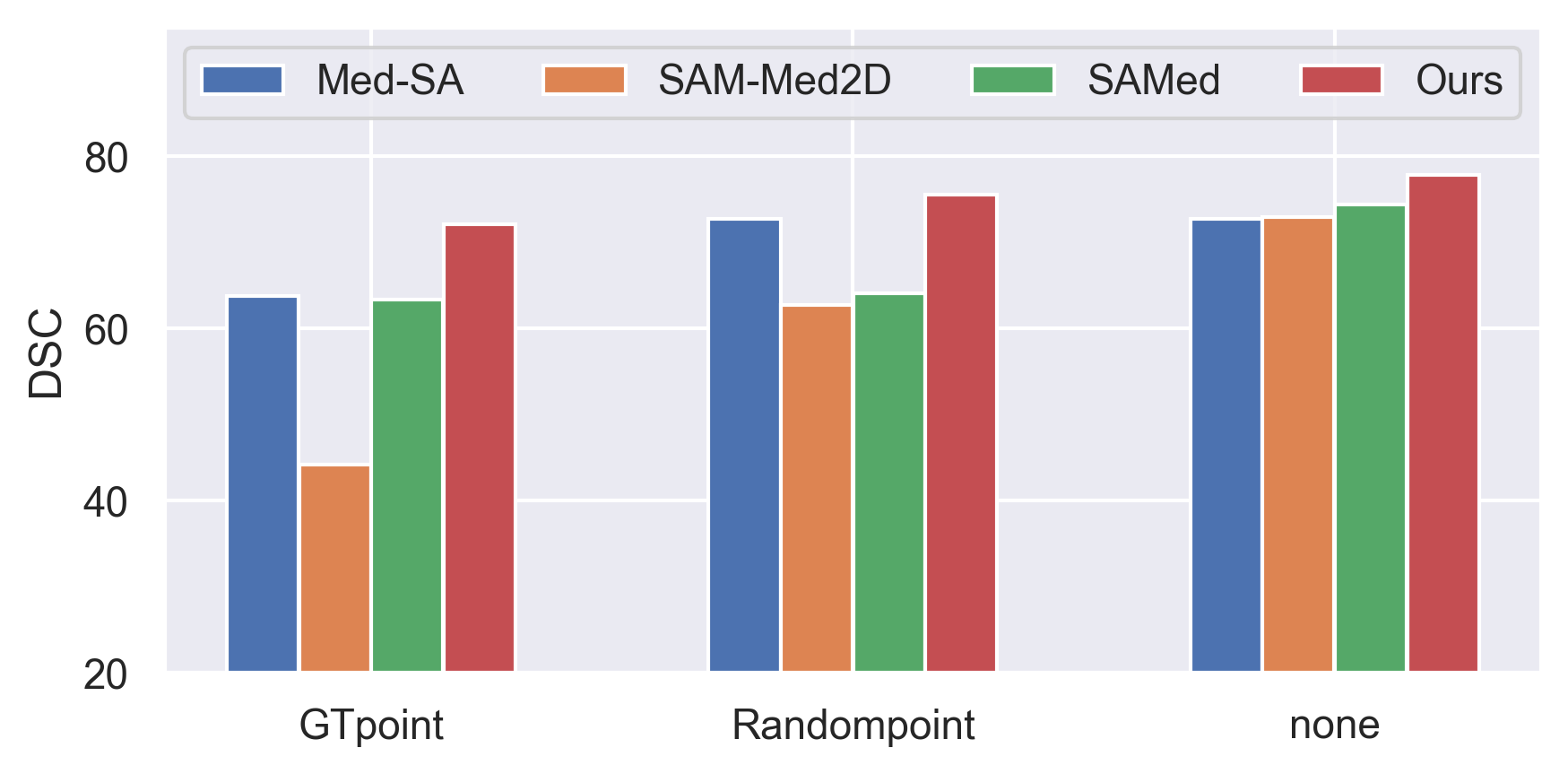}
    \caption{Comparison with different training strategies.}
    \label{fig:prompt} 
\end{figure}

As described previously, our SAM-SP and other SAM-based approaches do not use any prompts during both training and testing phases, achieving consistency between training and testing conditions.  To investigate  whether the different prompt strategies during training affect the inference performance or not. Following previous works \cite{ma2023segment, cheng2023sam}, we conduct evaluations on Seg-GPR with different choices of prompt strategies during training, including: no prompt (None), Ground-truth Random Point Prompt (Randompoint) and Ground-truth  Center Point Prompt (GTpoint). Notably, we still ensure that no prompts are provided during inference. As summarized in Fig \ref{fig:prompt}, SAM-SP, along with other SAM-based approaches, performs best when no prompts are used during training, compared to other training prompt strategies. Interestingly, although GTpoint provides more accurate ground-truth information, it leads to worse inference performance.
This counter-intuitive results may be due to that accurate prompts provided during training makes the model overly reliant on provided prompts, thus leading to significant performance degradation when no prompts are available during inference.
Additionally, the use of GTpoint prompts severely degrades the inference performance of SAM-Med2D. On the contrary, our SAM-SP experience only minor performance changes with GTpoint prompts. This may be attributed to  our proposed self-prompting module, which endeavor to produce reliable prompts autonomously, reducing reliance on externally provided prompts.

\section{Conclusion}
In this paper, we first emphasize the significance of reducing reliance on expert prompts when applying visual foundation models to specific domains, such as the Segment Anything Model (SAM), which broadens the application range of models. 
To alleviate the need for expert prompts in various specific domains, we introduce a novel self-prompting based fine-tuning approach SAM-SP, tailored for extending the capabilities of  the vanilla SAM model. Specifically, SAM-SP leverages its own output from the previous iteration as prompts to guide the subsequent iteration of the model. This self-prompting module endeavors to learn how to generate effective prompts autonomously and alleviates the dependency on expert prompts  during evaluation, significantly broadening SAM's applicability. Additionally,   we integrate a self-distillation module to further enhance the self-prompting process. Extensive experiments across various datasets validate the effectiveness of the proposed SAM-SP, which not only alleviates  the reliance on expert prompts but also exhibits satisfactory performance.
We hope that our study will benefit future research in self-prompting, and help reduce the dependence on expert prompts when applying visual foundation models to specific domains. 


\bibliography{aaai25}

\end{document}